\documentclass[a4paper]{./styles/svproc}
\usepackage{url}
\usepackage{graphics} 
\usepackage{epsfig} 
\usepackage{mathptmx} 
\usepackage{times} 
\usepackage{amsmath} 
\usepackage{amssymb}  
\usepackage{cite}
\usepackage{amsmath,amssymb,amsfonts}
\usepackage{algorithmic}
\usepackage{graphicx}
\usepackage{textcomp}
\usepackage{xcolor}
\usepackage{hyperref}
\usepackage{multirow}
\usepackage{multicol}
\usepackage{mathrsfs}
\usepackage{etoolbox}
\usepackage{amsmath}

\begin{document}
\mainmatter              
\title{DexGrasp-Diffusion: Diffusion-based Unified Functional Grasp Synthesis Method for Multi-Dexterous Robotic Hands}
\titlerunning{DexGrasp-Diffusion}  
%
\author{Zhengshen Zhang$^{1,*}$, Lei Zhou$^{1}$, Chenchen Liu$^{1}$, Zhiyang Liu$^{1}$, Chengran Yuan$^{1}$,
Sheng Guo$^{1}$, Ruiteng Zhao$^{1}$, Marcelo H. Ang Jr.$^{1}$, and Francis EH Tay$^{1}$
}
\authorrunning{Zhengshen Zhang et al.} 
%
\tocauthor{Zhengshen Zhang, Lei Zhou, Chenchen Liu, Zhiyang Liu, Chengran Yuan,
Sheng Guo, Ruiteng Zhao, Marcelo H. Ang Jr., and Francis EH Tay}

\institute{$^{1}$Advanced Robotics Centre, National University of Singapore, 117608, Singapore,\\
\email{robotics@nus.edu.sg},\\ WWW home page:
\texttt{https://arc.nus.edu.sg/}\\ $^{*}$Corresponding Author: \email{zhengshen\_zhang@u.nus.edu}
}

\maketitle              

\begin{abstract}
The versatility and adaptability of human grasping catalyze advancing dexterous robotic manipulation. While significant strides have been made in dexterous grasp generation, current research endeavors pivot towards optimizing object manipulation while ensuring functional integrity, emphasizing the synthesis of functional grasps following desired affordance instructions. This paper addresses the challenge of synthesizing functional grasps tailored to diverse dexterous robotic hands by proposing DexGrasp-Diffusion, an end-to-end modularized diffusion-based method. DexGrasp-Diffusion integrates MultiHandDiffuser, a novel unified data-driven diffusion model for multi-dexterous hands grasp estimation, with DexDiscriminator, which employs a Physics Discriminator and a Functional Discriminator with open-vocabulary setting to filter physically plausible functional grasps based on object affordances. The experimental evaluation conducted on the MultiDex dataset provides substantiating evidence supporting the superior performance of MultiHandDiffuser over the baseline model in terms of success rate, grasp diversity, and collision depth. Moreover, we demonstrate the capacity of DexGrasp-Diffusion to reliably generate functional grasps for household objects aligned with specific affordance instructions.
\end{abstract}

\section{Introduction}
The versatility of human grasping abilities is remarkable. In addition to the conventional five-fingered grasp, humans exhibit efficient generalization of grasping actions even under conditions where certain fingers are occupied \cite{li2023gendexgrasp}. Moreover, humans demonstrate an innate capacity to envision a diverse array of grasping configurations tailored to specific tasks, even when presented with different kinds of hands, achieving these adaptations rapidly and with a notable degree of success. In the area of robotics, the burgeoning interest in dexterous grasping stems from its ability to generate a diverse set of grasp candidates characterized by high success rates. Compared to parallel grippers, a primary advantage conferred by dexterous hands lies in their ability to firmly grasp and hold tools or other objects of diverse shapes and sizes \cite{bao2023dexart, turpin2022grasp} to facilitate the application of force \cite{agarwal2023dexterous}. So, it is more suitable and imperative to endow dexterous hands with the capability to perform functional grasps according to certain affordance instructions and utilize tools anthropomorphically.

\begin{figure}[htbp]
    \centering
    \includegraphics[width=0.8\linewidth]{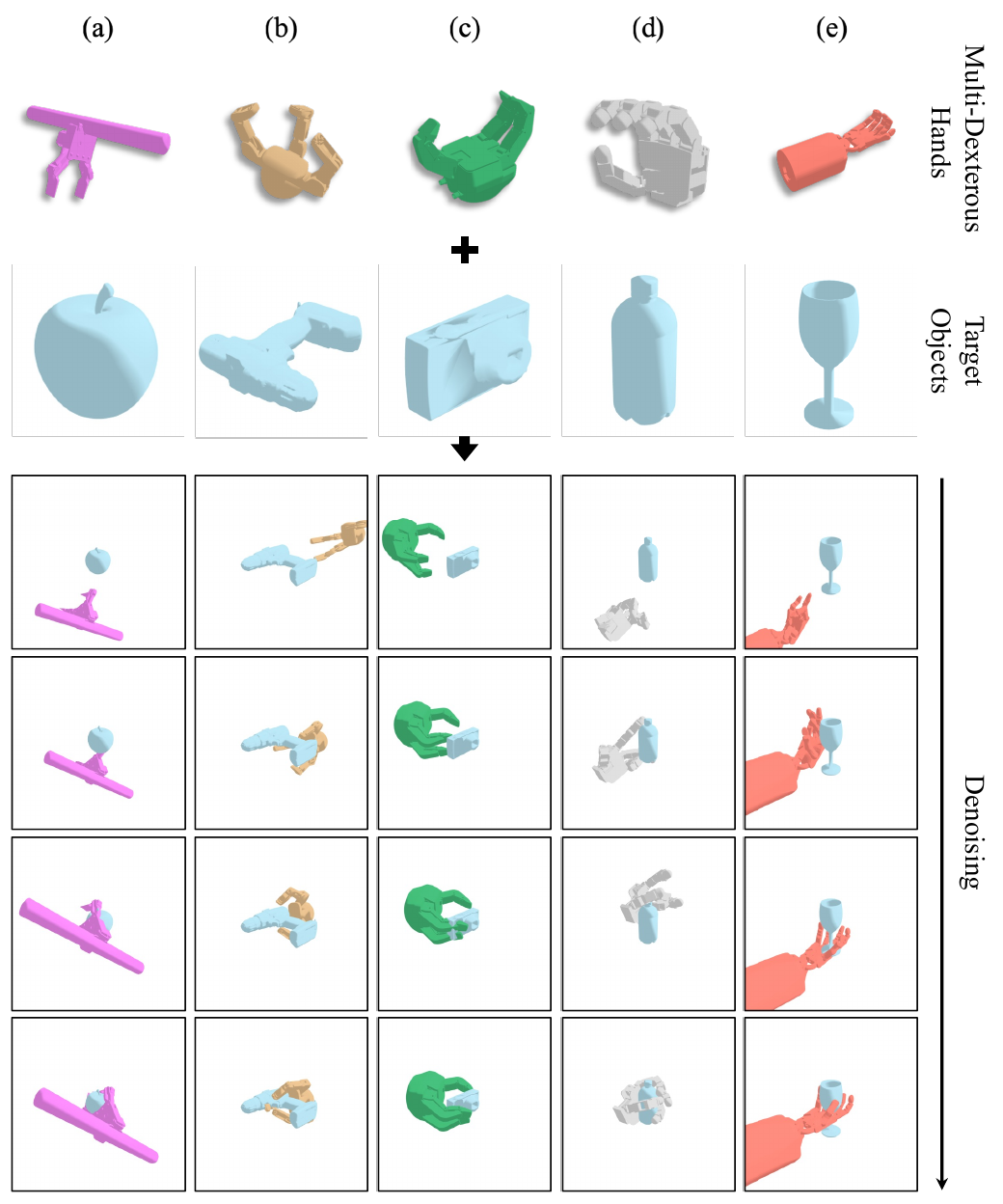}
    \caption{
       Overview of the diffusion sampling process with fixed target objects for multi-dexterous robotic hands performed by our presented DexGrasp-Diffusion method. (a) EZGripper with apple. (b) Barrett with power drill. (c) Robotiq-3F with camera. (d) Allegro with water bottle. (e) ShadowHand with wine glass.
    }
    \label{teaser}
    \vspace{-4mm}
\end{figure}


While prior research \cite{li2023gendexgrasp, huang2023diffusion} has demonstrated the ability to generate dexterous grasps, the ultimate objective for complex robotic manipulation tasks is to successfully grasp and utilize objects effectively. Hence, it becomes imperative for us to identify the object affordance regions, ensuring that the robot can grasp the object without impeding its intended functionality. Recent advancements in depth camera technology have spurred research efforts towards affordance detection in 3D point clouds \cite{9981900, deng20213d, nguyen2023open}, which most are approached as a supervised task involving labeling predetermined affordances for each point. Notably, \cite{nguyen2023open} introduced an innovative approach termed open-vocabulary affordance detection, diverging from predefined affordance labels by employing language models \cite{radford2021learning}. While this methodology enhances flexibility during affordance learning, it lacks the provision of grasp poses corresponding to the identified affordances. Consequently, the pursuit of universal affordance detection remains an exploration and poses challenges for practical implementation in robotic manipulation tasks. Some previous studies have merged affordance detection with grasp pose generation \cite{9981900, he2023pick2place}, yet they are constrained by predefined affordance sets and two-finger parallel grippers.

In response to the aforementioned challenges, we propose DexGrasp-Diffusion, an end-to-end modularized functional grasp synthesis method that combines multi-dexterous hand grasp estimation with open-vocabulary affordance detection to enhance the adaptability and manipulation abilities of dexterous hands for complex tasks (Fig.~\ref{teaser}). DexGrasp-Diffusion includes MultiHandDiffuser, a novel unified data-driven diffusion model that samples multi-dexterous hand grasps, and DexDiscriminator, which consists of a \textit{Physics Discriminator} and a \textit{Functional Discriminator}. Given an object 3D point cloud, our MultiHandDiffuser first generates diverse robust grasp poses of one specific hand. Thereafter, DexDiscriminator will eliminate physically invalid candidates and select suitable and feasible functional grasps associated with the desired affordances. We conduct experiments to assess DexGrasp-Diffusion's ability to generate physically plausible functional grasps on the MultiDex dataset \cite{li2023gendexgrasp}, which encompasses a varied array of grasp poses tailored for multiple dexterous robotic hands ranging from two to five fingers.  

Our main contributions are as follows:
\begin{enumerate}
    \item We propose a novel unified diffusion-based grasp generation network, MultiHandDiffuser, tailored for multiple dexterous robotic hands.
    \item We conduct comprehensive experiments on the MultiDex dataset to demonstrate that our MultiHandDiffuser outperforms the baseline model concerning success rate, diversity of sampled grasps, and collision depth.
    \item We integrate MultiHandDiffuser with DexDiscriminator and propose an effective modularized approach, DexGrasp-Diffusion, for generating feasible and reasonable functional grasps based on desired object affordances with open-vocabulary setting.
\end{enumerate}

\section{Related Works}

\subsection{Data-Driven Dexterous Grasping}
Prior to the emergence of large-scale datasets \cite{li2023gendexgrasp, wang2023dexgraspnet, 10160314}, considerable research efforts were devoted to exploring analytical and simulation-based approaches for multi-finger robotic hands grasping \cite{5654380, liu2021synthesizing, 1371616}, which are characterized by restricted generalizability or necessitate substantial computational resources. In contrast to the majority of analytical methodologies, data-driven methods evince enhanced inference speed and a broader spectrum of generated grasping configurations.
Data-driven methods for dexterous grasp synthesis can be broadly categorized into three primary approaches: 1) techniques that produce the object surface's contact map \cite{li2023gendexgrasp, wu2022learning}, 2) methodologies predicated upon shape completion principles \cite{lundell2021multi, van2020learning, lu2020multi, wei2022dvgg}, and 3) approaches that involve the training of grasping policies utilizing Reinforcement Learning algorithms \cite{mandikal2021learning} or human demonstration data \cite{wan2023unidexgrasp++}. However, a pervasive challenge encountered by many data-driven methods lies in reconciling the trade-off between diversity and grasp quality, and the diversity of produced grasps remains constrained by the composition and scope of the training dataset. In this work, the proposed method leverages the probabilistic nature and inherent randomness of the diffusion model to mitigate the limited diversity of sampled grasps to some extent while ensuring quality.




\subsection{Diffusion Models for Robotics}
Although still in its nascent stage, diffusion models have already demonstrated extensive utility within the field of robotics. Notable applications include manipulation \cite{chi2023diffusionpolicy, structdiffusion2023, simeonov2023rpdiff}, motion planning \cite{carvalho2023motion}, human-robot interaction \cite{ng2023diffusion}, and grasping \cite{urain2022se3dif, huang2023diffusion}. Among these, Urain et al. \cite{urain2022se3dif} presented a diffusion model trained to produce SE(3) grasp poses for parallel jaw. Huang et al. \cite{huang2023diffusion} introduced SceneDiffuser, a conditional diffusion-based model for various 3D scene understanding tasks and could synthesize stable and diverse grasp poses and human-like dexterous gripper configurations in all of SE(3). However, SceneDiffuser can only generate dexterous grasps for one specific dexterous hand but lacks the capability to learn and produce multi-dexterous hand grasps simultaneously. In contrast, our unified MultiHandDiffuser can generate complete and stable SE(3) grasp poses and hand configurations for five different dexterous hands and presents a superior performance than SceneDiffuser in benchmark testing.


Previous studies investigating diffusion models in robotic applications have additionally incorporated discriminators to assess the quality of the samples generated by the diffusion process \cite{structdiffusion2023, simeonov2023rpdiff}. For instance, in \cite{structdiffusion2023}, the discriminator is employed to evaluate the realism of point cloud scenes generated from the diffusion model, whereas in \cite{simeonov2023rpdiff}, the discriminator is tasked with assessing the effectiveness of a generated SE(3) pose for object manipulation. In this work, we combine two individual discriminators with MultiHandDiffuser to evaluate and select the physically valid functional dexterous grasp aligned with any unconstrained affordance label from the diffusion-produced grasp candidates.

\begin{figure*}[t]
    \centering
    \includegraphics[width=\linewidth]{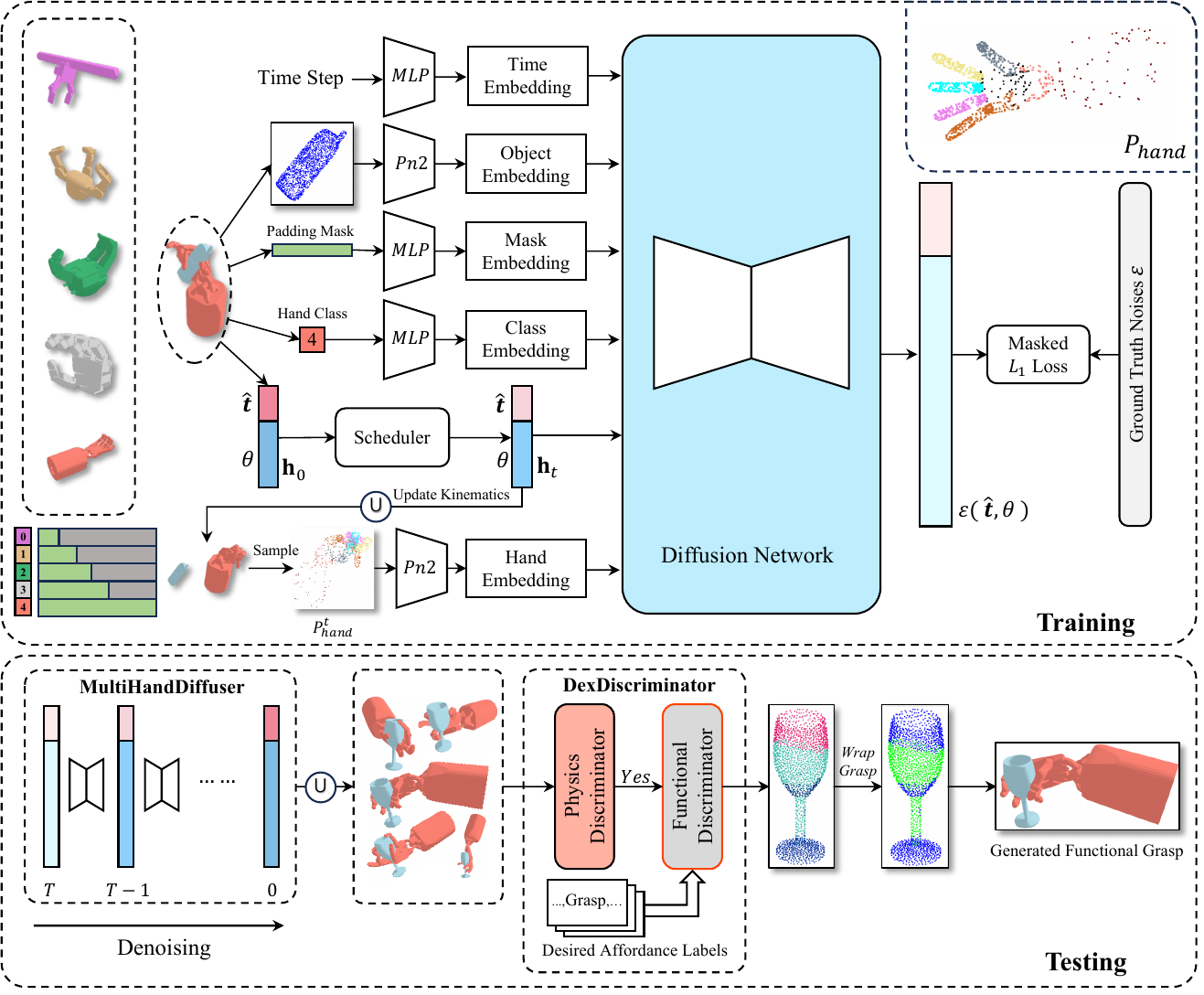}
    \caption{
      Overview of our proposed DexGrasp-Diffusion method.
    }
    \label{pipeline}
    \vspace{-5mm}
\end{figure*}

\section{Method}
\subsection{Problem Statement}
In this work, we assume that the representation of the given object \textbf{o $\in \mathbb{R}^{N \times 3}$ }is point cloud. Let \textbf{ h = (t, $\theta$)} represents a posture of the hand, where $\textbf{t} \in {\mathbb{R}}^{3}$ denotes global translation, $\theta \in \mathbb{R}^{k}$ represents the joint angles of the hand model, with $k$ denoting the degrees of freedom (DoF). Object point cloud \textbf{o} is first randomly rotated within its own frame, subsequently translation $\textbf{t}$ and joint angles $\theta$ are denoised with MultiHandDiffuser (Sec. \ref{MultiHandDiffuser}) to match the robotic hand with the rotated object point cloud, generating a diverse set $\mathbb{H} = {\{\textbf{h}_{i}}\}_{i=1}^{m}$ of hand postures within object frame. Afterwards, the generated hand postures are filtered by \textit{Physics Discriminator} (Sec. \ref{physics_dis}) for physically stable grasping postures and \textit{Functional Discriminator} (Sec. \ref{functional_dis}) for suitable functional grasps.

\subsection{MultiHandDiffuser}
\label{MultiHandDiffuser}
To reduce training difficulty, we follow SceneDiffuser to transform the training data from  \textbf{ h = (t, R, $\theta$)} to  \textbf{ h = (t, $\theta$)}, by rotating the object point cloud and hand pose with $\textbf{R}^{-1}$, where $\textbf{R} \in {\mathbb{R}}^{3 \times 3}$ represents the rotation of robotic hand in object frame. In this way, the objective of our MultiHandDiffuser is to denoise translation $\textbf{t}$ and joint angles $\theta$ of the hand until it satisfies the desired geometric relationship with the rotated object point cloud. Afterwards, by rotating object point cloud and sampled hand pose with $\textbf{R}$, an object-centric grasp pose representation can be obtained, ensuring diversity of sampled grasp poses while simplifying the training process.

In order to handle varying lengths of $\theta$ due to different DoF of each hand, the length of $\textbf{h}$ is fixed to be 27, as ShadowHand has the most DoF (24). For the rest hands, $\theta$ is padded with 0 for invalid joints.
To develop a unified model that can generalize to multiple dexterous robotic hands, several conditions (as shown in Fig.~\ref{pipeline}) are applied to the diffusion process to generate diverse and high-quality grasp poses for the individual hands.

\textbf{Time Embedding.} For each time step $t$, it is encoded by a multilayer perceptron (MLP) to obtain time embedding. 

\textbf{Object Embedding.} To extract comprehensive object geometry features from point cloud \textbf{o} for feature matching and grasp generation, PointNet++ \cite{qi2017pointnet++} is employed as feature extractor to obtain object embedding. 

\textbf{Mask Embedding.} 
With the purpose of developing a unified model for multiple dexterous hands, our MultiHandDiffuser takes the $\textbf{h}_{t}$ of fixed length as input for training and denoising. However, simply padding invalid joint parameters of all hands except ShadowHand with $0$ and computing loss for all of the joints during training is prone to confuse the network. To mitigate this issue, we introduce a padding mask $M$ of equivalent dimensionality to $\textbf{h}$. For translation and valid joint parameters, the corresponding values in the padding mask are set to be $1$, while the values of padded joints are set to be $0$. Subsequently, padding mask $M$ is encoded as mask embedding and fused with hand pose features to further inform our network which joints are valid.


\textbf{Class Embedding.} To further differentiate between individual dexterous hands, each hand is labeled by a hand class $c = 0, \dots,4$, which is encoded into a feature vector and further encoded by an MLP layer to get class embedding. 

\textbf{Hand Embedding.}
Besides, for each time step $t$, the kinematics of the hand is updated with $\textbf{h}_{t}$, and subsequently, the hand point cloud is sampled from the current hand mesh model. The sampled hand point cloud $P_{hand} \in \mathbb{R}^{N \times 4}$ is accompanied by finger labels in the fourth column as an additional feature to further differentiate the point cloud of different fingers. To be more specific, points of palm and individual fingers are labeled from $0$ to $8$, which is visualized as a colored point cloud in the upper-right corner of Fig. \ref{pipeline}. Afterwards, hand point cloud $P_{hand}$ is encoded by another PointNet++ module to extract the hand's semantic information as well as geometric features for better matching between object and hand.

In the MultiHandDiffuser, mask embedding and class embedding are early fused with hand pose features to inform the network with hand type. Then, cross-attention is performed between the hand pose feature and the object embedding, as well as between the hand pose feature and the hand embedding to further guide the hand pose denoising process with comprehensive geometric and semantic information.

\textbf{Training:} 
The training process (forward) is a pre-defined discrete-time Markov chain in the hand pose space $\mathbb{H}$ spanning all possible hand poses represented as $\textbf{h}$. Given a ground truth hand pose $\textbf{h}_{0}$ in the dataset, Gaussian noise $\epsilon$ is gradually added to $\textbf{h}_{0}$ to obtain a series of intermediate hand poses $\textbf{h}_{1}$,...,$\textbf{h}_{T}$ with same dimensionality as $\textbf{h}_{0}$, according to a pre-defined noise scheduler. The diffusion model predicts the noise added to $\textbf{h}_{0}$ as $\epsilon(\hat{\textbf{t}}, \theta)$, subsequently masked L1 loss is computed as:

\begin{equation}
   \mathcal{L} = M  |\epsilon(\hat{\textbf{t}}, \theta) - \epsilon|.
\end{equation}

\textbf{Inference:}
Given a noisy hand pose sampled from a standard multivariate Gaussian distribution $\textbf{h}_{T} \sim \mathcal{N}(\textbf{0}, \textbf{I})$ as the initial state, it corrects $\textbf{h}_{t}$ to less noisy pose $\textbf{h}_{t-1}$ at each time step by the trained MultiHandDiffuser model with aforementioned conditions. By repeating this reverse process until the maximum number of steps $T$, we can reach the final state $\textbf{h}_{0}$, which is the grasp pose we aim to obtain.

\subsection{Physics Discriminator}
\label{physics_dis}
MultiHandDiffuser demonstrates proficiency in generating a diverse array of dexterous grasp candidates. However, it is important to recognize that a proportion of the produced grasp candidates may not lead to successful grasping outcomes. In order to filter out those physically invalid candidates and select the reasonable functional grasps aligned with any unconstrained object affordances, we introduce DexDiscriminator, which consists of two discriminators (Fig.~\ref{pipeline}) to assess all of the samples generated by MultiHandDiffuser.

We first validate whether a grasp belongs to a physically plausible grasp using the Isaac Gym environment \cite{makoviychuk2021isaac}, which is equipped with the basic physics engine PhysX. The validation is conducted by subjecting the object to external acceleration and observing its resultant movement. Each grasp undergoes testing involving the application of a uniform $ 0.5ms^{-2} $ acceleration to the object for a duration of 1 second, equivalent to 60 simulation steps. We ascertain the success of a grasp by evaluating whether the object displaces more than 2cm when the simulation ends. This testing procedure is repeated six times, with accelerations applied along the $x$, $y$, and $z$ axes. A grasp is deemed unsuccessful if it fails in any of the six tests. We also implement a refinement process informed by contact awareness for all sampled grasps across all dexterous hands, as diffusion models often manifest minor inaccuracies leading to instances of penetration or floating around contact regions. Initially, a goal pose is established by adjusting the joint links to positions in close proximity (within 5mm) to the object and oriented toward its direction. Following this, the pose parameter vector $ \textbf{h} $ undergoes an update via the gradient descent with a single step, directed towards mitigating the disparity between the present and goal poses. Eventually, the adjusted pose is monitored through the utilization of a positional controller integrated within the Isaac Gym.

\subsection{Functional Discriminator}
\label{functional_dis}
The role of the \textit{Functional Discriminator} here is to choose reasonable functional dexterous grasps guided by desired affordance instructions among all physically feasible grasp candidates. Concretely, we follow the recent open-vocabulary 3D point cloud affordance detection method, OpenAD \cite{nguyen2023open}, to detect the desired affordance region on an object specified by the text (as shown in Fig. \ref{pipeline}). Initially, an object's full point cloud \textbf{o $\in \mathbb{R}^{N \times 3}$ }is utilized as input for a PointNet++ model to systematically derive $ z $ pointwise feature vectors denoted as $ \textbf{C}_1 $, $ \textbf{C}_2 $, …, $ \textbf{C}_z $. Subsequently, the $ n $ linguistic labels associated with desired affordances undergo processing through a freeze CLIP \cite{radford2021learning} text encoder $ \chi $ to produce $ n $ word embeddings, denoted as $ \textbf{e}_1 $, $ \textbf{e}_2 $, …, $ \textbf{e}_n $. Then, in order to facilitate open-vocabulary affordance detection, we ascertain the semantic associations between the affordance descriptors of the point cloud and their prospective labels through the correlation of word embeddings and pointwise features using the cosine similarity function. Specifically, the correlation score denoted as $ S_{x,y} $, located at the intersection of the $ x $-th row and $ y $-th column within the correlation matrix $ \textbf{S} \in \mathbb{R}^{z \times n} $, represents the degree of correlation between the point feature $ \textbf{C}_x $ and the affordance word embedding $ \textbf{e}_y $. $ S_{x,y} $ is calculated as:

\begin{equation}
S_{x,y} = {\frac{\textbf{C}_x^\top \textbf{e}_y}{\|\textbf{C}_x\| \|\textbf{e}_y\|}}.
\end{equation}
The softmax outcome for an individual point $ x $ is calculated in accordance with the following expression:

\begin{equation}
O_{x,y} = {\frac{\exp\left(S_{x,y}/{\upsilon}\right)}{\sum_{m=1}^{n} \exp\left(S_{x,m}/{\upsilon}\right)}},
\end{equation}
the parameter $ \upsilon $ is subject to learning. This calculation is performed individually for each point within the object point cloud \textbf{o} to obtain the affordance label for every point. Subsequently, upon selecting a specific affordance label, the point cloud $\textbf{P}_{aff}$ corresponding to this particular affordance is retained, while extraneous points are filtered out.



For every grasp within the $ k $ grasp candidates that successfully pass the \textit{Physics Discriminator}, we label the contact region between the hand surface and the object surface point by computing the aligned distance \cite{li2023gendexgrasp} between them, thus acquiring a set of point clouds $ \textbf{P}_{1}^{con} $, $ \textbf{P}_{2}^{con} $, …, $ \textbf{P}_{k}^{con} $ restricted to points located within each corresponding contact region. Then, the Chamfer distance (CD) $ d_{CD}(\textbf{P}_{i}^{con}, \textbf{P}_{aff}) $ is individually computed between every contact region point cloud $ \textbf{P}_{i}^{con} $ and the object's affordance area point cloud $ \textbf{P}_{aff} $ using the following formula \cite{fan2017point}:

\begin{equation}
\begin{split}
d_{CD}(\textbf{P}_{i}^{con}, \textbf{P}_{aff}) & = \frac{1}{\textbf{P}_{i}^{con}} \sum_{p \in \textbf{P}_{i}^{con}} \min_{q \in \textbf{P}_{aff}} \| p - q \|_2^2 \\ 
& + \frac{1}{\textbf{P}_{aff}} \sum_{q \in \textbf{P}_{aff}} \min_{p \in \textbf{P}_{i}^{con}} \| q - p \|_2^2 .
\end{split}
\end{equation}
Ultimately, the grasp associated with the minimum CD $\min \limits_{i \in [1, k]}{d_{CD}(\textbf{P}_{i}^{con}, \textbf{P}_{aff})} $ is chosen as the most suitable functional dexterous grasp corresponding to the particular affordance label of the object.

\section{Experiments}
In this section, we undertake a series of experiments aimed at elucidating the efficacy of our proposed DexGrasp-Diffusion approach on the MultiDex dataset. Initially, we commence by comparing our approach with SceneDiffuser baseline models, which are exclusively trained using single-hand data. Subsequently, we furnish diverse ablation studies to facilitate a comprehensive investigation of the MultiHandDiffuser. Thirdly, we showcase noteworthy qualitative outcomes attained through DexGrasp-Diffusion. Lastly, an analysis of failure instances and prospective research directions is deliberated.

\subsection{Dataset}
In this work, MultiDex dataset \cite{li2023gendexgrasp} is used for training and testing, which contains five subsets (EZGripper, Barrett Hand, Robotiq-3F, Allegro, ShadowHand) of diverse dexterous grasping poses with 58 daily objects. Following SceneDiffuser \cite{huang2023diffusion}, the dataset is split into a training set (48 objects) and a testing set (10 objects) respectively.

\subsection{Evaluation Metrics}
 We conduct a set of quantitative assessments of DexGrasp-Diffusion, evaluating its performance based on metrics encompassing success rate, diversity, and collision depth. \textbf{Success Rate:} We evaluate the success of a grasp within the Isaac Gym by subjecting the object to external forces and subsequently measuring its displacement. \textbf{Diversity:} The assessment of grasp diversity entails the computation of standard deviation across joint angles among grasps that successfully pass the Isaac Gym test. \textbf{Collision Depth:} We quantify collision depth as the maximum penetration depth of the hand into the object during every successful grasp, serving as a metric to assess the performance of models in achieving physically valid grasps. 
 

\subsection{Implementation Details}
We train the MultiHandDiffuser for noise prediction through optimization using the Adam optimizer, employing a learning rate of 1e-4. Default values are retained for other Adam hyperparameters. Training the MultiHandDiffuser spans 2000 epochs with the batch size of 64 on a single NVIDIA 3090Ti GPU.

\section{Results and Analysis}
\subsection{Quantitative Analysis}
\textbf{Baselines.} We adopt SceneDiffuser as our baseline model. Since it is designed for grasp generating with single-hand model, we trained SceneDiffuser on each subset of the MultiDex dataset and obtained five distinct models to handle each hand model. Our proposed MultiHandDiffuser is directly trained on the Multidex dataset for all five dexterous hands. The evaluation result, as shown in Tab. \ref{tabresults}, demonstrates that our model achieves higher mean accuracy, higher grasp diversity, and less collision compared to baseline models. This superior performance not only ensures better overall efficacy but also underscores the model's enhanced capability to generalize across multi-dexterous hands. It is noteworthy that the CAD model of the Robotiq-3F from the MultiDex dataset experiences severe self-collisions within the Isaac Gym environment, which cause inaccurate evaluation data with the \textit{Physics Discriminator} as these collisions displace objects. Thus, results related to the Robotiq-3F are excluded from our analysis.

\begin{table}[!h]
		\caption{A comparison between different models on the MultiDex dataset. \textbf{Succ.}: Success Rate (\%)$\uparrow$. \textbf{Div.}: Diversity (rad.)$\uparrow$. \textbf{Col.}: Collision Depth (mm)$\downarrow$. $\uparrow$: The higher, the better. $\downarrow$: The lower, the better. \underline{ }: Indicates the second-best result.}
        \vspace{-5mm}
		\label{tabresults}
        \scriptsize
\begin{center}
  
  \begin{tabular}{p{1.4cm}|p{0.62cm}|p{0.62cm}|p{0.62cm}|p{0.62cm}|p{0.62cm}|p{0.62cm}|p{0.62cm}|p{0.62cm}|p{0.62cm}|p{0.62cm}|p{0.62cm}|p{0.62cm}|p{0.62cm}|p{0.62cm}|p{0.62cm}}
			\hline
			\hline
			\multirow{2}*{Model} & \multicolumn{3}{c|}{Ezgripper} & \multicolumn{3}{c|}{Barrett} & \multicolumn{3}{c|}{Allegro} &  \multicolumn{3}{c|}{Shadowhand} &  \multicolumn{3}{c}{Mean}\\
            \cline{2-16}
			& {\textbf{Succ.}} & {\textbf{Div.}} & {\textbf{Col.}} & {\textbf{Succ.}} & {\textbf{Div.}} & {\textbf{Col.}} & {\textbf{Succ.}} & {\textbf{Div.}} & {\textbf{Col.}} & {\textbf{Succ.}} & {\textbf{Div.}} & {\textbf{Col.}} & {\textbf{Succ.}} & {\textbf{Div.}} & {\textbf{Col.}}\\
            \hline
            baseline & 26.41& 0.181& 18.11& 20.78& 0.233& 15.42& \textbf{40.00}& 0.142& \textbf{16.70}& 63.59& 0.158& 18.57& 37.70& 0.179& 17.20 \\
            handclass & 44.69& 0.170& 14.84& \textbf{27.34}& 0.213& 14.12& 32.50& 0.181& 17.26& 61.72& 0.171& \textbf{15.34}& 41.56 & 0.184& \textbf{15.39}\\
            pc & 39.84& 0.176 & 17.45 & 19.38& 0.251& 15.62& 38.44 & 0.205 & 19.35 & 63.13 & 0.213 & 17.28 & 40.20 & 0.211& 17.43 \\
            pc+handclass & 45.78 & 0.166 & 15.67 & 21.72 & 0.246 & 14.49 & 33.75 & \textbf{0.226} & 18.85 & 67.03 & \textbf{0.225} & 18.36 & \underline{42.07} & \textbf{0.216} & 18.84\\
            fullset & \textbf{49.22}& \textbf{0.206} & \textbf{13.68}& 25.78& \textbf{0.258} & \textbf{13.18}& 35.94& 0.196 & 17.10& \textbf{67.97}& 0.196& 17.67& \textbf{44.73} & \underline{0.214}& \underline{15.41}\\
      

			\hline
            \hline
		\end{tabular}
	\end{center}
  \vspace{-4mm}
\end{table}




\textbf{Ablation Study.}
To evaluate the effects of different conditions on our MultiHandDiffuser, we conduct thorough ablation studies by removing each condition from the network input. Through those ablation studies, in order to use a unified model for all hands, padding mask is kept to inform our network which are valid joints. The first configuration is denoted as \textit{handclass}, which is most similar to the baseline except we fuse class embedding into the hand pose features to inform which hand the network is dealing with. Intuitively, the baseline model is separately trained on each subset, which is supposed to outperform \textit{handclass} model trained on the whole MultiDex dataset spanning across five hands. However, \textit{handclass} outperforms the baseline model in all three aspects (mean success rate, diversity, and collision depth). We suspect that the training samples of each subset are few, resulting in overfitting of the baseline model to the training set. Besides, training across all five subsets enables the network to learn how to match each hand to the object's geometry instead of memorizing the hand pose in the training set.

The second set is denoted as \textit{pc}, which means we discard hand class and finger label and only extract hand embedding on unlabeled hand point cloud. Performance on both success rate and collision drops as the network is no longer explicitly informed which type of hand it is dealing with, making it more challenging for both training and inference.
The third set is denoted as \textit{pc+handclass}, which represents using unlabeled hand point cloud as condition while informing the network which hand it is handling. Results in Tab. \ref{tabresults} show that it not only outperforms \textit{pc} but also outperforms \textit{handclass}, meaning that the combination of hand point cloud and hand class enables the network to take advantage of diverse training samples while capturing the geometric features of the hand for better matching between objects and hands.  

In \textit{fullset}, all of the conditions including finger label are provided to the network, further enabling it to differentiate between multiple hands and different fingers. Therefore, \textit{fullset} achieves the highest mean success rate, second-best and comparable diversity and collision depth. 
Furthermore, we systematically vary the diffusion time steps $T$ for \textit{fullset} model and report the mean value of each metric in Tab. \ref{tabresults2}. We observe that $T$ plays a crucial role in balancing the diversity and success rate of dexterous grasp estimation. Specifically, a value of $T$ = 100 yields optimal diversity in produced grasps, while $T$ = 1000 results in the highest mean success rate.

 \begin{table}[!h]
	\renewcommand\arraystretch{1.0}
        \setlength\tabcolsep{10pt}
		\caption{A comparison between different \textit{fullest} models on the MultiDex dataset. \textbf{Succ.}: Success Rate (\%). \textbf{Div.}: Diversity (rad.). \textbf{Col.}: Collision Depth (mm). $\uparrow$: The higher, the better. $\downarrow$: The lower, the better.}
        \vspace{-4mm}
		\label{tabresults2}
\begin{center}

     
     
     

 \begin{tabular}{p{1.5cm}|p{0.60cm}|p{0.60cm}|p{0.60cm}}
            \hline
            \hline
            \multirow{2}{*}{Model} & \multicolumn{3}{c}{Mean} \\
            \cline{2-4}
            & \textbf{Succ.}$\uparrow$ & \textbf{Div.}$\uparrow$ & \textbf{Col.}$\downarrow$ \\
            \hline
            fullset\textsubscript{100} & 41.33 & \textbf{0.262} & 16.42 \\
            fullset\textsubscript{500} & 44.26 & 0.230 & 16.42 \\
            fullset\textsubscript{1000} & \textbf{44.73} & 0.214 & \textbf{15.41} \\
            \hline
            \hline
        \end{tabular}
    \end{center}
\end{table}

\subsection{Qualitative Results}

\begin{figure*}[!h]
    \centering
    \includegraphics[width=\linewidth]{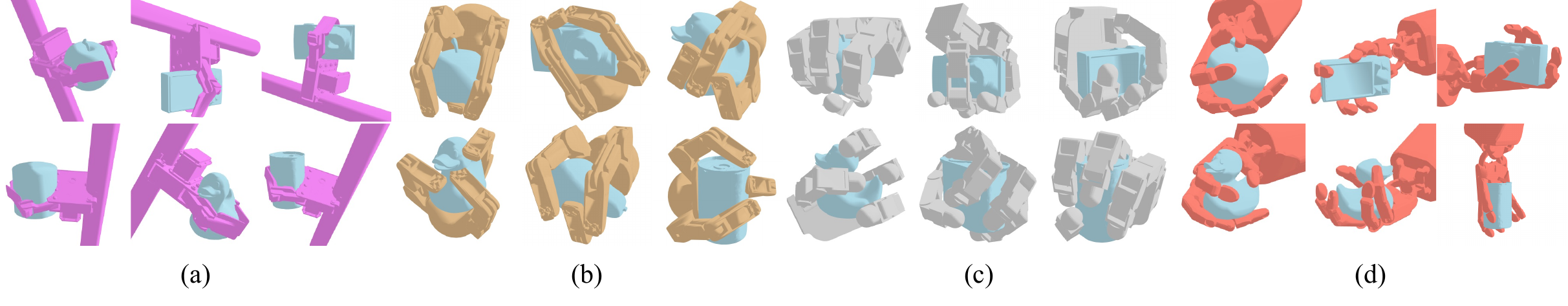}
    \caption{
      Generated physically plausible grasp candidates for unseen objects. (a) EZGripper. (b) Barrett. (c) Allegro. (d) ShadowHand.
    }
    \label{result1}
    \vspace{-5mm}
\end{figure*}



\textbf{Generalization to Unseen Objects.} Fig. \ref{result1} presents diverse high-quality outcomes produced by DexGrasp-Diffusion on the testing set of the MultiDex dataset. The generated grasps exhibit a wide range of diverse grasping modalities, including but not limited to hooks, squeezes, wraps, tripods, and other variations. Moreover, grounded in the assurance of diversity, each grasp candidate has undergone meticulous scrutiny by the \textit{Physics Discriminator} to ensure physical plausibility, thereby furnishing a robust repository of samples for subsequent assessment by the \textit{Functional Discriminator}.

\begin{figure}[!h]
    \centering
    \includegraphics[width=0.85\linewidth]{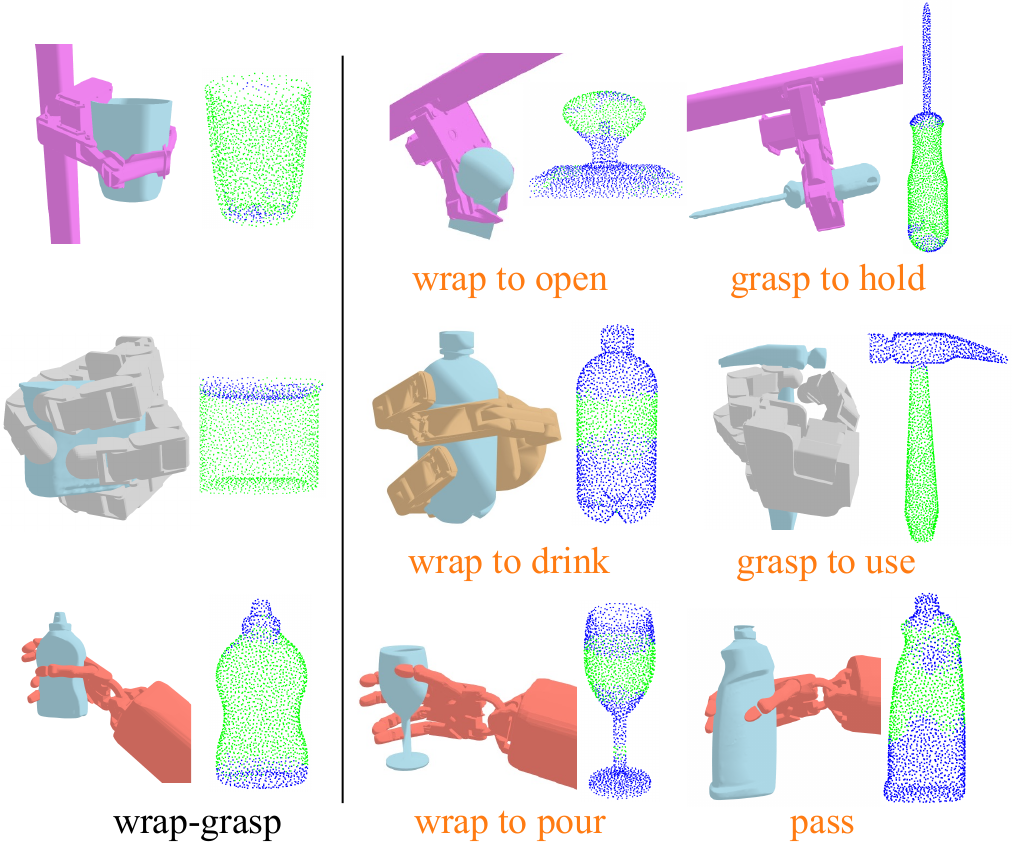}
    \caption{
      Qualitative results of detected functional grasps by DexGrasp-Diffusion. The unseen affordances are shown in \textcolor{orange}{orange}.
    }
    \label{result2}
    \vspace{-4mm}
\end{figure}

\textbf{Generated Functional Grasps.} In Fig. \ref{result2}, we present several illustrative instances utilizing objects sourced from the MultiDex dataset, which effectively showcase our pipeline's capacity to generate functional grasps aligned with desired affordance labels. Notably, for simple and seen affordances in the training set of OpenAD \cite{nguyen2023open}, the attainment of high-quality affordance detection outcomes substantially facilitates the selection of dependable functional grasps by DexGrasp-Diffusion. Conversely, when confronted with challenging, previously unseen affordance labels, although the derived affordance regions may not exhibit absolute precision, our DexGrasp-Diffusion algorithm adeptly discerns and filters rational and suitable functional grasps, which proves the robustness of the proposed method and its ability to handle different desired affordance labels of varying complexity.

\subsection{Discussion}

\begin{figure}[!h]
    \centering
    \includegraphics[width=0.9\linewidth]{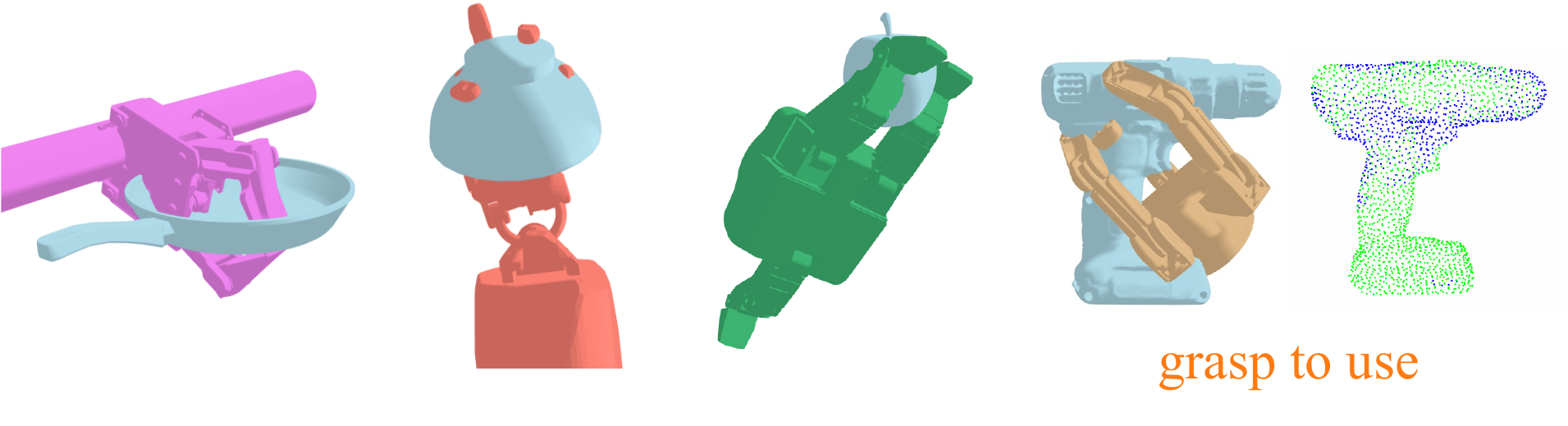}
    \caption{
      Some failure or counter-intuitive cases of our method. The unseen affordances are shown in \textcolor{orange}{orange}.
    }
    \label{result3}
    \vspace{-3mm}
\end{figure}

Despite yielding promising results, it is imperative to acknowledge that our method has not achieved flawless proficiency in multi-dexterous hand grasp synthesis and universal functional grasp detection. Instances, where DexGrasp-Diffusion manifests its limitations, are delineated in Fig. \ref{result3}. Specifically, the first two cases depict instances wherein our MultiHandDiffuser failed to generate viable grasps for the bowl and pan, despite attempts by the hands to access and grasp the bottom regions of said objects. We believe that these failure examples stem from the absence of explicit collision constraints during the training phase of MultiHandDiffuser, leading to instances where the hand disregards potential collisions and proceeds to grasp the wrong positions. Furthermore, the scarcity of objects with similar shapes within the training dataset contributes to diminished success rates in MultiHandDiffuser's grasp predictions for objects such as bowls and pans. Consequently, having a comprehensive largescale multi-dexterous hand dataset, encompassing objects with more intricate geometries would facilitate enhanced model training. Subsequently, an interesting yet counter-intuitive grasp is observed in the third case depicted in Fig. \ref{result3}, wherein the model endeavors to grasp an apple with two fingers while extending another finger towards the opposing end. We attribute this phenomenon to analogous physically feasible but unreasonable ground truth grasp poses within the MultiDex dataset, thereby resulting in the learning of corresponding data distributions during the training of MultiHandDiffuser. 

The final case exemplifies an occurrence where our approach generates a false-positive grasp, which fails to be a desired functional grasp according to the given affordance label due to the inaccurate and noisy affordance detection by \textit{Functional Discriminator}. Furthermore, we also note that only the orientation of the input object point cloud is consistent with the orientation of the object in the OpenAD's training set, the OpenAD model may obtain appropriate affordance detection results, which weakens the robustness of our method to a certain extent. Given the modular nature of our method, substituting OpenAD with a more adaptable and stable open-vocabulary 3D point cloud affordance detection module could ameliorate this limitation.

\section{Conclusions}
In summary, we propose DexGrasp-Diffusion, an innovative unified framework for generating physically feasible functional grasp poses tailored to multi-dexterous robotic hands.
This method effectively addresses prior limitations by seamlessly integrating diffusion-based grasp synthesis with open-vocabulary affordance detection into the dexterous functional grasp generation process. 
Experimental evaluation conducted on the MultiDex dataset substantiates the superior performance of DexGrasp-Diffusion in terms of success rate, grasp diversity, and collision depth when compared to baseline models, and its capability of producing viable and reasonable functional grasps for household objects guided by desired affordances.
We hope that the outcomes of DexGrasp-Diffusion will motivate future researchers to advance robotic manipulation, leading to the development of more intelligent and autonomous robotic systems that could better understand and perform in complex environments.

\bibliographystyle{IEEEtran}
\bibliography{root}

\end{document}